\documentclass[10pt,twocolumn,letterpaper]{article}

\usepackage{cvpr}
\usepackage{times}
\usepackage{epsfig}
\usepackage{graphicx}
\usepackage{amsmath}
\usepackage{amssymb}
\usepackage{algorithm}
\usepackage{algpseudocode}
\usepackage{amsmath}
\usepackage{graphics}
\usepackage{epsfig}
\usepackage[square, comma, sort&compress, numbers]{natbib}
\usepackage{graphicx}
\usepackage[utf8]{inputenc}
\usepackage{mathtools}
\usepackage[table,xcdraw]{xcolor}

\usepackage[normalem]{ulem}
\useunder{\uline}{\ul}{}

\usepackage[breaklinks=true,bookmarks=false]{hyperref}

\cvprfinalcopy 

\def\cvprPaperID{****} 

\begin{document}

\title{ Rethinking the constraints of multimodal fusion:  case study in Weakly-Supervised Audio-Visual Video Parsing}

\author{Jianning Wu \small{$^{1,b}$} \quad\large{ Zhuqing Jiang\small{$^{1,2,a,b}$}} \quad Shiping Wen\small{$^3$} \quad \large{Aidong Men\small{$^1$} }\quad Haiying Wang\small{$^1$}\\
\small{$^a$} \small Corresponding author\\
\small{$^b$} \small The first two authors contribute equally to this work\\
\small{$^1$} \small School of Artificial Intelligence, Beijing University of Posts and Telecommunications, Beijing, China\\
 \small{$^2$} \small Beijing Key Laboratory of Network System and Network Culture, Beijing University of Posts and Telecommunications, Beijing, China\\
 \small{$^3$} \small Australian AI Institute, Faculty of Engineering and Information Technology, University of Technology Sydney, Australia\\
{\tt\small $\{$jianningwu,jiangzhuqing$\}$@bupt.edu.cn
}
}
\maketitle


\begin{abstract}
   For multimodal tasks, a good feature extraction network should extract information as much as possible and ensure that the extracted feature embedding and other modal feature embedding have an excellent mutual understanding. The latter is often more critical in feature fusion than the former. Therefore, selecting the optimal feature extraction network collocation is a very important subproblem in multimodal tasks. Most of the existing studies ignore this problem or adopt an ergodic approach. This problem is modeled as an optimization problem in this paper. A novel method is proposed to convert the optimization problem into an issue of comparative upper bounds by referring to the general practice of extreme value conversion in mathematics. Compared with the traditional method, it reduces the time cost. 

Meanwhile, aiming at the common problem that the feature similarity and the feature semantic similarity are not aligned in the multimodal time-series problem, we refer to the idea of contrast learning and propose a multimodal time-series contrastive loss(MTSC). 

Based on the above issues, We demonstrated the feasibility of our approach in the audio-visual video parsing task. Substantial analyses verify that our methods promote the fusion of different modal features.

\end{abstract}

\section{Introduction}
Audio-Visual Video Parsing(AVVP)\cite{tian2020unified} has a wide potential application in downstream video understanding tasks.(such as monitoring analysis, video summarization, and retrieval).It is a newly introduced multi-modal task that involves detecting and localizing occurrences of events within the audio and visual streams of a video. \begin{figure}[h!]
\centering
\includegraphics[scale=0.4]{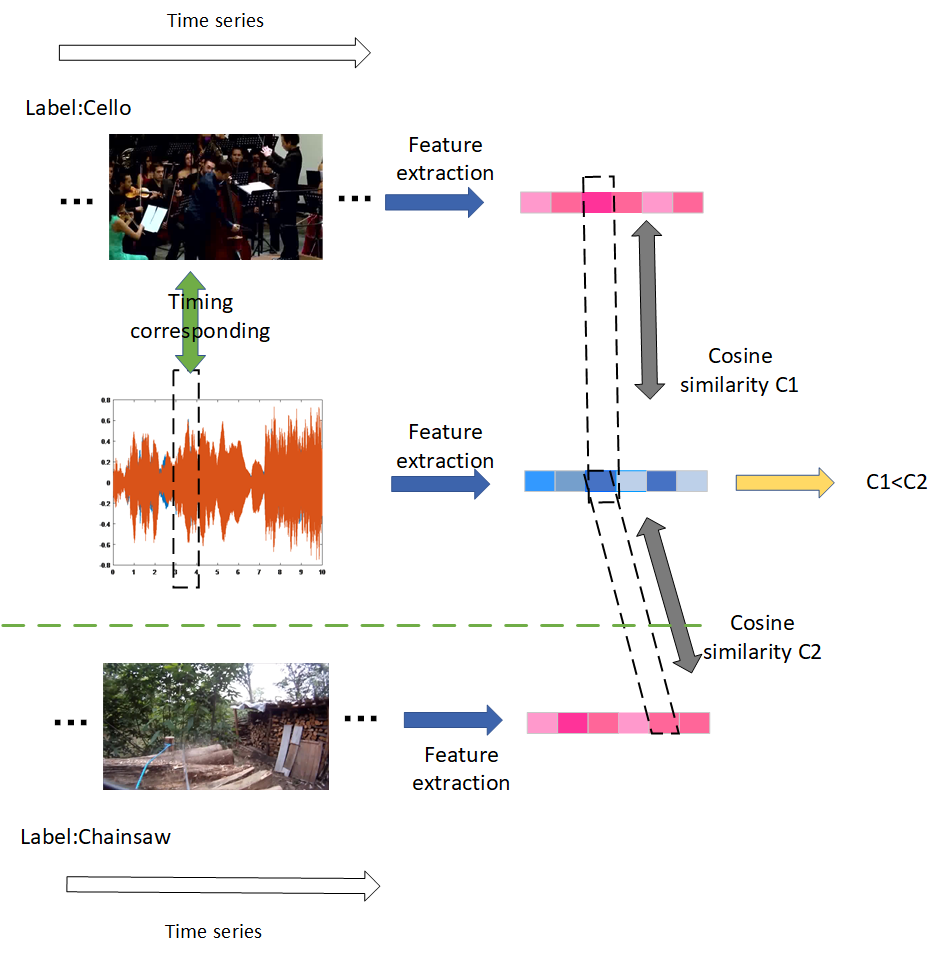}
\caption{C2 is the maximum similarity of the audio fragments in the figure.C1 is the similarity between the image frame of the corresponding segment and the audio at the same time; C2 is the similarity between the image frames and the audio of different video clips.}
\label{fig:example}
\end{figure}It directly contributes to audio-visual source separation, especially when the sources of audio are occluded in the video. 

\begin{figure*}[h!]
\begin{center}
\includegraphics[scale=0.3]{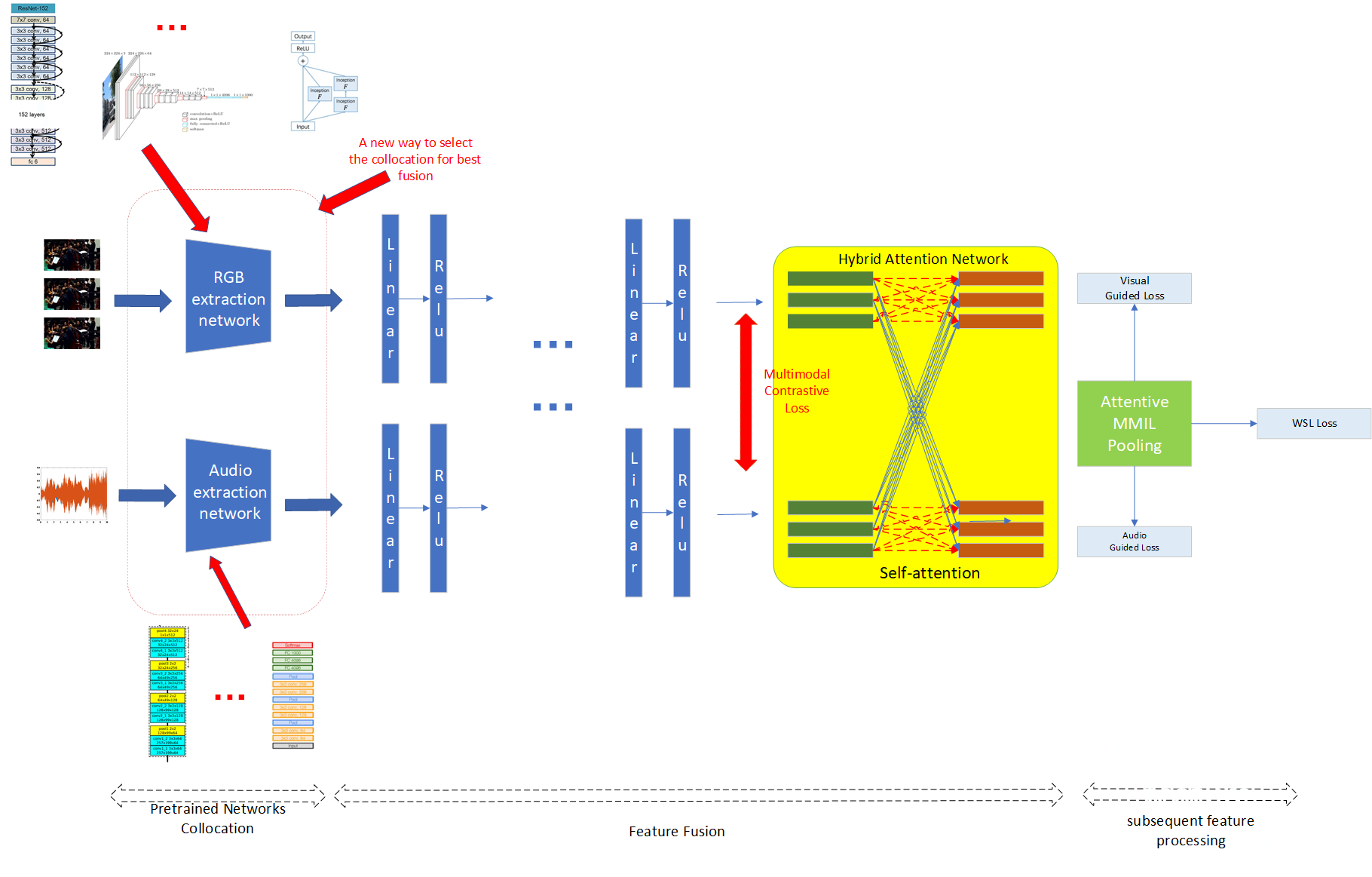}
\caption{Our methods on baseline}
\label{fig:system}
\end{center}
\end{figure*}

In general, the standard process of the multimodal problem\cite{sun2019videobert,yang2020bert,su2019vl,li2021scheduled,huang2020m3p} could be divided into three steps: feature extraction/embedding, feature fusion, and subsequent feature processing. Among them, the most characteristic and challenging step for multimodal missions is multimodal fusion\cite{baltruvsaitis2018multimodal}.(As shown in Figure \ref{fig:system}) Obtained through the first two parts, the fusion feature plays an import role in the working performance of subsequent feature processing.

However, when designing network structure and loss function, the existing AVVP method\cite{tian2020unified} only takes the characteristics of TAL task into consideration and pays less attention to the difficulty and complexity of multimodal feature fusion. Considering it, we propose that the main AVVP methods have two problems:

1.Lack of a practical approach to select pre-trained feature extraction network collocation effectively.

2.Ignoring the difference between feature similarity and feature semantic similarity.

The first problem exists not only in only AVVP but also in most of the multimodal tasks as long as the extraction network is needed. The inevitability of this problem is mainly based on the following fact: 
1. Except for the difference inside the raw multi-mode data, the feature extraction network also significantly impacts subsequent feature fusion, which is ignored and underestimated in common sense; 
2. there are plenty of pre-trained feature extraction networks but few methods for selecting them, except traverse; 
3. Meanwhile, in today's most popular multimodal presentation learning task, the cost of a pure training time has reached kTPU*day(Tensor Processing Unit)\cite{su2019vl}. Such a traversal method is not acceptable as the datasets expand and the participated modes increase. Therefore, an effective alternative method is needed to reduce the time cost and promote feature fusion.

In practice, We model the problem as an optimization problem. (As shown in Eq.\ref{euq2}) However, due to the limitation of the gradient descent, it is challenging to ensure getting the extreme point\cite{bottou2012stochastic,kingma2014adam}. Since this training method takes a long time and it is difficult to ensure the results obtained, we turn to look for the upper bound of this extreme value and verify through experiments that our method is  positively correlated to the traditional method. Meanwhile, our method greatly reduces the selecting time.(Meanwhile, this method proposed by us reduces the time needed for selection.)

In the step of multimodal feature fusion, the existing multimodal WS-TAL task\cite{tian2020unified,lin2019dual,tian2019audio,tian2018audio,wu2019dual} mainly uses various Transformer variants, such as Hybrid Attention Network(HAN), to fuse features, which implicitly assumes that feature similarity is equivalent to feature semantic similarity. This assumption is valid for single modes, but the differences of the similarities between modes are much greater in multi-mode problems.

To verify above judgment, the similarities within and between the extracted audio features and video features are calculated on the LLP dataset(Look, Listen, and Parse)\cite{tian2020unified}. Within one mode, 97 \% of the two features with the maximum similarity were labeled identically,while between modes, this rate drop to less than 30 \%. (See Table \ref{tab1} for details) The probability reveals that there is a severe difference between feature similarity and semantic similarity in multimodal tasks.( Figure \ref{fig:example} shows one example  ) Therefore, It is inappropriate to utilize the multimodal Transformer structure directly for single-mode tasks, which will greatly damage the performance of feature fusion. Moreover, semantic constraints and corrections should be completed befor inter-modal features before using them.

Also, it should be noted that, the difference in image features between adjacent frames is slight. However,as for TAL domain problems, most of the valuable information is hidden in these differences. The AVVP task will be adequately addressed if we can extract the information.

In light of these, referring to the idea of contrastive learning: reduce the distance between positive samples and increase the distance between negative samples, we introduce a multimodal contrastive loss. Due to the lack of segment level labels, we consider two modal features at the same time as positive samples, and those at different time as negative samples. It immensely enhances the correlation between feature similarity and feature semantic similarity among various modes, which is of great benefit for feature fusion. In this respect, Zhang\cite{zhang2021cola} is similar to us. The differences between him and us lie in: 1. In his work, contrastive learning is introduced to solve the unimodal WS-TAL problem. He believes that(His reason is that) indistinguishable snippets could be easily misclassified and hurt the localization. 2. Different from him, we introduce our method in the multimodal task to solve the disalignment between the feature similarity and feature semantic similarity.

The main contributions of this paper are:

1. Pioneeringly,in the step of feauture embedding, considering that feature extraction plays a vital role in feature fusion in multimodal tasks, we propose an effective method to select the collocation of different model feature extraction networks. Compared with the traversal method currently used, our method ensures effectiveness and reduces time complexity.

2. In the step of feature fusion, a multi-mode time-series contrastive loss(MTSC) is proposed to promote feature fusion between modes given the disalignment of feature similarity and semantic similarity in AVVP. It also helps to the expression of the difference information.

3. Both methods we proposed helps to facilitate the feature fusion. We demonstrated the feasibility of our method in the specific AVVP task. However,the two problems we proposed are common across the multimodal tasks. The solution can be generalized to other multimodal task.


\section{Related works}
1.\textbf{WS-TAL}: Weakly supervised temporal action localization only requires video-level annotation, which has attracted wide attention.UntrimmedNets\cite{wang2017untrimmednets} solve this problem by first classifying clip suggestions and then selecting relevant clips in a soft or hard manner.STPN\cite{nguyen2018weakly} imposes a sparsity constraint to enforce the sparsity of the selected segment. Hide-and-seek \cite{singh2017hide}and MAAN\cite{yuan2019marginalized} extend the identification area by randomly hiding plaques or inhibiting the dominant response, respectively. Zhong et al.\cite{zhong2018step} introduced a progressive generation program to achieve a similar purpose. W-TALC\cite{paul2018w} applies depth measurement learning to complement the multi-instance learning formula. Unlike actions in TAL, video
events in audio-visual video parsing might contain motionless or even out-of-screen sound sources and the events can be perceived by either audio or visual modalities.
.

2.\textbf{Contrastive Learning}: Contrastive Learning is a kind of self-supervised learning, which aims to learn knowledge by oneself from unlabeled data instead of relying on labeled data.
Contrastive means learning to use internal data patterns to learn an embedding space in which correlated signals are aggregated, and non-correlated signals are distinguished by Noise Contrast Estimation (NCE)\cite{gutmann2010noise}.CMC(Contrastive Multiview Cocoding)\cite{tian2019contrastive} proposes a Contrastive learning framework that maximizes the mutual information between different views of the same scenario to achieve view-invariant representation.SIMCLR\cite{chen2020simple} selects the negative sample by using an enhanced view of the other items in the small batch.
Compared with the positive samples of comparison learning under the single mode, which also need to be obtained through data enhancement (SIMCLR) or different views of the same scene (CMC), timing multimode problems naturally have a pair of positive samples of different modes, and the information between different modes is supplementary rather than complementary. This naturally creates a unique application advantage for the application of comparative learning in multimode. To our knowledge, this paper is the first to combine contrastive learning with time-series information in multimodal TAL related tasks. The experimental results show that the contrast loss introduced in this paper is beneficial to multimodal time series localization.

3.\textbf{Multimodal feature fusion}: Fusion is a key research topic in multimodal studies, which integrates information extracted from different unimodal data sources into a single compact multimodal representation. There are three types of methods that are mainly used to fuse audio with image features, namely, simple operation-based\cite{zoph2016neural,anastasopoulos2019neural,nojavanasghari2016deep}, attention-based\cite{li2019visualbert,sun2019learning,cho2020x,shih2016look,bahdanau2014neural,yang2016stacked,xu2016ask,anderson2018bottom}, and tensor-based methods\cite{tenenbaum2000separating,zadeh2017tensor,gao2016compact}. Among them, attention-based approach is the main research direction. However, attention needs the high correlation between feature similarity and feature semantic similarity, which is invalid between modes. Therefore, a novel multimodal time-series contrastive loss is proposed by us to enhance the correlation.

4.\textbf{Pre-trained feature extraction networks}: For convenience, there are many pre-trained extraction networks\cite{he2016deep,simonyan2014very,hu2018squeeze,zhang2017polynet,zhang2020split,phan2017improved,cramer2019look,arandjelovic2017look} for future embedding. The backbones and train sets of them vary, while all of the networks get a remarkable performance in multiple tasks after fine-tuning. However, in multimodal tasks, besides the remarkable performance of a single network,
the mutual "understanding" between features in different modes extracted from the varied networks matters. We demonstrate experimentally that the latter is even more important than the former in multimodal tasks. Therefore, the collocation of pre-trained feature extraction networks with different modes is of great research value. Based on this, this paper proposes a novel method.
\section{Approach}

In this section, we introdeuce our method in detail.First, we explain some of the high frequency words(Section 3.1). Then, we describe our proposed method. In order, it consists two main parts: (i) a novel way to select the feature extraction networks(section 3.2).
(ii) a multimodal time-series contrastive loss for better alignment.(section 3.3)

\subsection{Explaination of key words }
First of all, we explain some key words we often use.Like \cite{zhang2020multimodal},we use the word "networks" to refer to pre-trained feature extraction networks; When we talk about AVVP's task network, we use the word "model".We use the words "mode" and "multimodal" respectively to describe the concepts of single and cross-modal.

\subsection{A novel method for the collcation of pretrained feature extraction networks }
\subsubsection{Task objective function}
Assume that there are N modes, and each mode has M pre-trained extraction network selections. Then the task's optimization goal is

\begin{equation}
\mathop {\arg }\limits_{{\Phi _{_1}}, \cdots ,{\Phi _{\rm{N}}}} \mathop {\max }\limits_E E(\Phi _1^{{j_1}}({x_1}), \cdots ,\Phi _{\rm{i}}^{{j_{\rm{i}}}}({x_1}), \cdots ,\Phi _{\rm{N}}^{{j_{\rm{N}}}}({x_1}),\theta )
\end{equation}
Here,${\Phi _i}$represents ith mode and ${\Phi ^{^{{j_i}}}}$is any chioce of one of the M networks in the ith mode, $\mathrm{E}$ is the target function,$\theta$ means parameters of the model.

Assume that each training time is $T_0$ and the inference time is $T_1$, then the time to traversal all cases is $MN*\left( {{T_0} + {T_{1}}} \right)$. Of course, the demand will consume more time in the actual situation because of time costs to adjust hyperparameters and other reasons, but it should be of this order.

The existing selection method is mainly traversal, and it is mathematically formulated as followed.

\begin{equation}
\max _{E} \mathrm{E}\left(\phi\left(\mathrm{A}_{2}\right), \theta\right)\mathop {,\arg }\limits_{\phi ,\theta } \mathop {\max }\limits_E E(\phi_i ({A_{_1}}),\theta )
\label{euq2}
\end{equation}

where ${\phi _i} \in \{ \Phi _1^{{j_1}}({x_1}),\Phi _2^{{j_2}}({x_1}), \cdots ,\Phi _{\rm{i}}^{{j_{\rm{i}}}}({x_1}), \cdots ,\Phi _{\rm{N}}^{{j_{\rm{N}}}}({x_1})\} $

However, this method has the disadvantages of a long time consuming and many traversal numbers.

\subsubsection{A feasible solution}

Since the training time cost is much longer than the test time cost, the most effective way to reduce the cost is to reduce the training time. Furthermore, due to the limitation of gradient descent, it is impossible to traverse every point on the loss hyperplane. So it is difficult to verify whether a comparison obtained by the above method is the same as the actual case (or just the result of inadequate tuning). In view of the above two purposes, we change the comparison object. To overcome the handicap of calculating the above Max formula's exact value, we turn to find an upper bound of the value, which is much easier to realize. Thus, we manage to change the comparative upper bound to measure the selection schemes quickly and effectively.

The traditional method is to find the optimal solution by training in the training set and testing in the test set.It directly calculates the optimal solution of all the specific tasks corresponding to the network, and then selects the optimal solution. This method is actually a joint solution of selecting the optimal model collocation and finding its. If we can skip the problem of finding the optimal model collocation result, choose the optimal solution in another way, and then calculate the performance of the optimal solution, then the costs can be reduced to O(T0+T1). For multi-modal problems, this substitution method exists:
\begin{equation}
\begin{split}
\left\lceil {\mathop {\max }\limits_E {\rm{E(}}\phi {\rm{(}}{{\rm{A}}_2}{\rm{),}}\theta {{\rm{)}}_{\mathop {\arg }\limits_{\phi ,\theta } \mathop {\min }\limits_L L(\phi ({A_1}),\theta )}}} \right\rceil \\
 = \mathop {\max }\limits_E {\rm{E(}}\phi {\rm{(}}{{\rm{A}}_2}{\rm{),}}\theta {{\rm{)}}_{\mathop {\arg }\limits_{\phi ,\theta } \mathop {\max }\limits_E E(\phi ({A_2}),\theta )}}
\end{split}
\end{equation}
where$\left\lceil  f \right\rceil $ represents the  upper bound on the function ${f}$;${A_2}$presents test set;${A_1}$presents trainning set and eval set.;$\mathop {\max }\limits_E {\rm{E(}}\phi {\rm{(}}{{\rm{A}}_2}{\rm{),}}\theta {{\rm{)}}_{\mathop {\arg }\limits_{\phi ,\theta } \mathop {\max }\limits_E E(\phi ({A_1}),\theta )}}$represents the process of training on the training set and optimizing on the eval set

A reasonable explanation is that: a model trained on the training set normally performs worse than one trained on the test set when both are tested on the test set -- because the latter will have serious overfitting.Nevertheless, the overfitting is what we wanted. The more overfitting   a collocation has, the easier it catch the multimodal informations.Therefore, the latter is sufficient to compare the effects of different feature network options on feature fusion.In this way, we successfully separated the selection of the appropriate feature network collocation from the training of the entire network on the training set, reducing the time to O(T1).

We regard the test result trained on the test set as F1, and the result trained on the training set as F2.The only drawback of the above argument is that it's hard to ensure  the trend of change in F1 scores can reveal the trend of F2 when the collocation changes. In the subsequent experiments, we have proved that the change rule of F1 and F2 is roughly the same under various collocations of the mainstream feature extraction network. Such a result is understandable because the difference is mainly related to the subsequent model, whereas this model is fixed in our experiment.

\subsection{MTSC}

In view of the mismatch between feature similarity and feature semantic similarity in Multimodal tasks, as well as the fact that useful information is not effectively extracted in some modes, we refer to the idea of SIMCLR\cite{chen2020simple} and introduce Multimodal Time-series Contrastive Loss(MTSC).

\begin{figure*}[htbp]
\centering
\begin{minipage}[t]{0.48\textwidth}
\centering
\includegraphics[width=6cm]{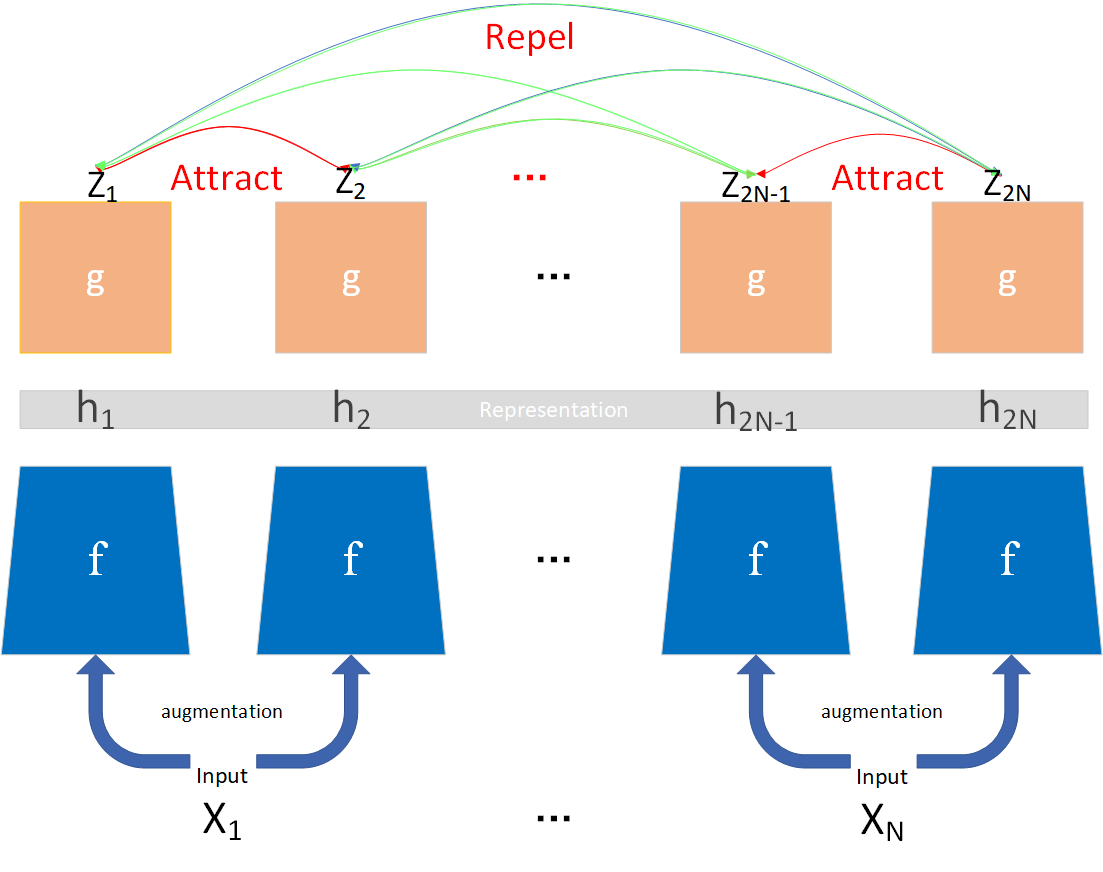}
\caption{SimCLR\cite{chen2020simple}}
\label{fig:simclr}
\end{minipage}
\begin{minipage}[t]{0.48\textwidth}
\centering
\includegraphics[width=6cm]{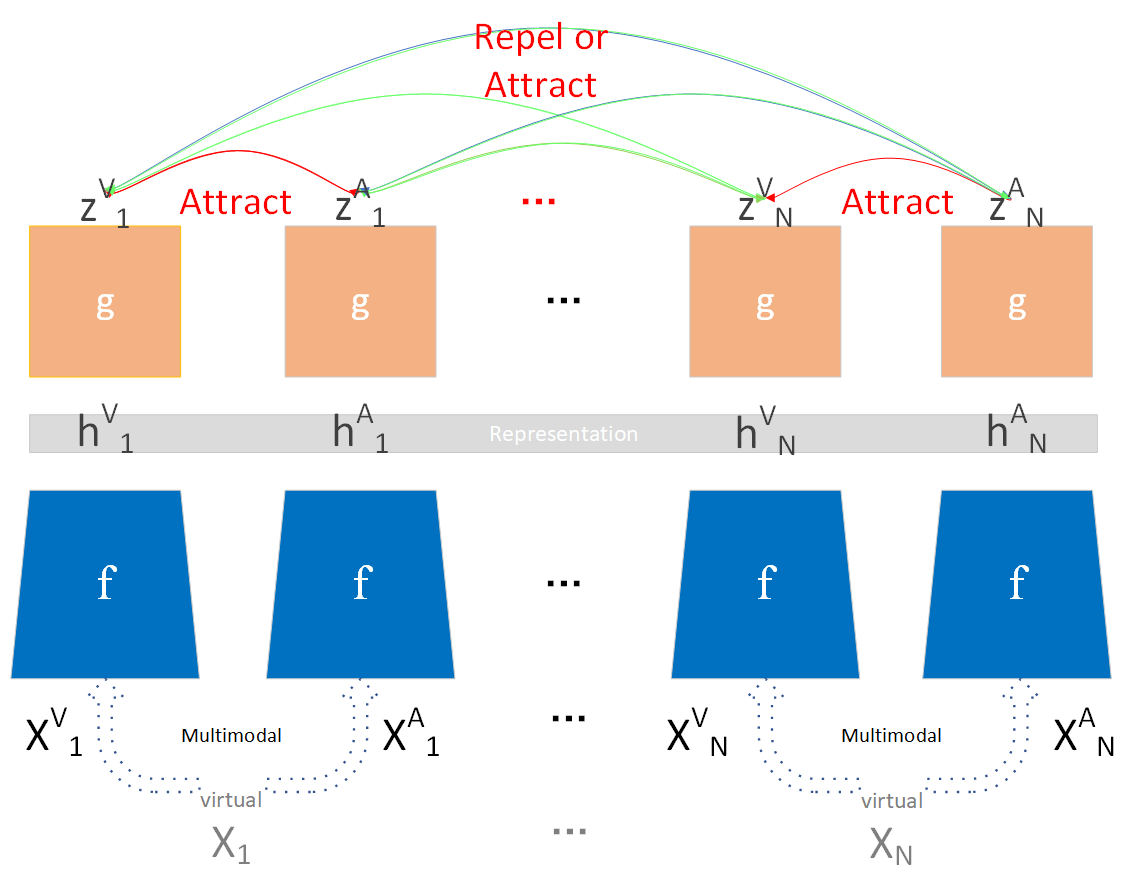}
\caption{Multimodal SimCLR}
\label{fig:sim_multi}
\end{minipage}
\end{figure*}

In SIMCLR\cite{chen2020simple}(As shown in Figure.\ref{fig:simclr}), for  \emph{N} samples in a minibatch, SIMCLR adopts two different data enhancement methods to obtain  \emph{2N} sample points. SimCLR does not clearly indicate the negative sample pair. On the contrary, for a pair of positive sample points enhanced from a single data, The remaining  \emph{2(N-1)} enhancement samples in minibatch are regarded as negative samples.

\begin{equation}
l_{i}=-\log \frac{\exp \left(\operatorname{sim}\left(z_{i}^{1}, z_{i}^{2}\right) / \tau\right)}{\sum_{k=1}^{N} 1_{[k \neq i]} \exp \left( \operatorname{sim}\left(z_{i}^{1}, z_{\mathrm{k}}^{2}\right) / \tau\right)}
\end{equation}

Different from the simple discrimination of positive and negative samples in SIMCLR, we have a priori information of time series in this task, that is, based on such a fact : semantic correlation decreases with the increase of time difference, so our loss function can be optimized as

\begin{equation}
\begin{split}
l_{i}=-\log \frac{\exp \left(\operatorname{sim}\left(\phi_{i}^{1}, \phi_{i}^{2}\right) / \tau\right)}{\sum_{k=1}^{N} 1_{[k \neq i]} \exp \left(T^{*} \operatorname{sim}\left(\phi_{i}^{1}, \phi_{\mathrm{k}}^{2}\right) / \tau\right)},\\
 T=F(\mathrm{k}-i)
\end{split}
\label{eq5}
\end{equation}

Due to the insensitivity of Eq.\ref{eq5} to the exponential part, the equation is modified as:

\begin{equation}
\begin{split}
l=\frac{1}{N^{2}} \sum_{j=1}^{N} \sum_{i=1}^{N}\left[sim \left(\phi_{i}^{1}, \phi_{j}^{2}\right)-F(j-i) \cdot S_{i, j}\right]\\
S_{i, j}=T_{i, j} \cdot sim_{detach} \left(\phi_{i}^{1}, \phi_{j}^{1}\right)
\end{split}
\label{euq:ours}
\end{equation}
where $T_{i,j}$ is a time lag coefficient which increases as the time lag decreases;$sim_{detach}$ represents a detach of$sim$.We set  $T_{i,j}={\rm{1/}}{\left| {i - j} \right|^5}$. 

Compared with the SIMCLR task, in the AVVP task, it is difficult to determine whether the features at different moments comprise a positive sample pair or a negative sample pair(As shown in Figure.\ref{fig:sim_multi}). Therefore, we choose to use the relationship between features in a single mode to make adaptive discrimination, namely $sim_{detach}$ in the Eq.\ref{euq:ours}.

We use time series information to increase the similarity of features of different modes, and reduce the similarity of features of different modes. While narrowing the gap between feature similarity and feature semantic similarity, the information contained in the feature difference value is re-extracted.

\begin{algorithm}[h] 
\begin{algorithmic}[1]

\Require
batch size $N$,structure of $f,t,g,\varphi  $ $f$ is the feature extraction networks;$t$ is part of the model before HAN; $g$ is a Linear layer.
\For {sampled minbatch$\{ {\textbf{x}_k}\} _{k = 1}^N$}
\ForAll{$k \in \{ 1, \ldots ,N\} $}
draw two modes $\Phi\sim \varphi,\Phi ^{'}\sim \varphi $
\State $//$the first mode
\State${{\widetilde x_k} } = {f_1}(\Phi ({x_{_k}}))$
\State${h_k} = t({\widetilde x_k})$
\State${z_k} = g({h_k})$
\State $//$the second mode
\State${{\widetilde x_k}^{'} } = {f_2}(\Phi^{'} ({x_{_k}}))$
\State${h_k^{'}} = t({\widetilde x_k}^{'})$
\State${z_k^{'}} = g({h_k}^{'})$
\EndFor
\ForAll {$i \in \{ 1, \ldots ,N\} $ and $j \in \{ 1, \ldots ,N\} $}
\State ${s_{i,j}} = {z_i}z_j^{'}/(\left\| {{z_i}} \right\|\left\| {z_j^{'}} \right\|)$
\EndFor
\State\textbf{define} ${l_{i,j}}$ \textbf{as} Eq. \ref{euq:ours}
\State $L = {\textstyle{1 \over N^{2}}}\sum\nolimits_{j = 1}^N\sum\nolimits_{i = 1}^N  {l(i,j)} $
\State update t,g to minimize $L$ 
\EndFor

\Return model $t( \cdot )$,and throw away $g( \cdot )$
\caption{ MTSC's main learning algorithm}
\label{alth1} 
\end{algorithmic} 
\end{algorithm}
Like SimCLR, Algorithm  \ref{alth1} summarizes the proposed method.

In the specific application, to retain more semantic integrity of feature,  Algorithm \ref{alth1} could be integrated into the model training. To alleviate the problem of under-fitting states of different modes in the original model, we preserve the g(·) in visual brach. 

\section{Experiments}
\subsection{Implementation Details}
 We consider the same experimental settings as Tian did. For a 10-second-long video, we first sample video frames at 8fps, and each video is divided into non-overlapping snippets of the same length with eight frames in one second.Batch size and epochs are 16 and 40. The initial learning rate is 3E-4 and will drop by multiplying 0.1 after every ten epochs. Our models optimized by Adam can be trained using one NVIDIA 2080 GPU.

\textbf{Dataset} we evaluate our method on the LLP datasets which is designed for the AVVP task. LLP contains 11,849 YouTube video clips spanning over 25 categories for a total of 32.9 hours collected from AudioSet \cite{gemmeke2017audio}. A wide range of video events (e.g., human speaking, singing, baby crying, dog barking, violin playing, and car running, and vacuum cleaning etc.) from diverse domains (e.g., human activities, animal activities, music performances, vehicle sounds, and domestic environments) are included in the dataset.

\textbf{Pretrained feature extraction networks} We use two famous audio feature extraction networks including vggish\cite{gemmeke2017audio} and openl3-audio\cite{cramer2019look},and five popular RGB image feature extraction networks including resnet-152\cite{he2016deep},vgg19bn\cite{simonyan2014very},polynet\cite{zhang2017polynet},SENET-154\cite{hu2018squeeze} and openl3-image\cite{cramer2019look}.

\textbf{Baselines} We compare our method with the existing AVVP method\cite{tian2020unified} proposed by Tian. Since our contribution does not involve network design, our experiment is carried out on the network of Baseline.(As shown in Figure \ref{fig:system})

\textbf{Evaluation Metrics} As Tian\cite{tian2020unified} did, we evaluate them on parsing all types of events (individual audio, visual, and audio-visual events) under both segment-level and event-level metrics. To evaluate overall audio-visual scene parsing performance, we also compute aggregated results, where Type@AV computes averaged audio, visual, and audio- visual event evaluation results and Event@AV computes the F-score considering all audio and visual events for each sample rather than directly averaging results from dierent event types as the Type@AV. We use both segment-level and event-level F-scores \cite{mesaros2016metrics} as metrics. The segment-level metric can evaluate snippet-wise event labeling performance. We extract events with concatenating consecutive positive snippets in the same event categories and compute the event-level F-score based on mIoU = 0.5 as the threshold for computing event-level F-score results. Besides, Considering that this is a Multimodal Multiple Instance Learning(MMIL) problem with both segment-level task and event-level task, we consider a new Average Scores averaging the segments' and events' scores mentioned above as one of the final criteria for evaluating the performance of the model.

\subsection{Experimental Comparison}
\subsubsection{the disalignment between feature semantic similarity and feature similarity  in different modes}

First, we verify a severe misalignment of the existing model between different modes, that is, the misalignment between feature semantic similarity and feature similarity. The details are shown in Table \ref{tab1}. Specifically, within one mode, 97.14 \% of the two features with the maximum similarity were labeled identically, while between modes, this rate dropped to 29.18 \%.

Then, the matching degree of feature semantic similarity and feature similarity between modes is compared in the feature fusion stage with/without the contrastive loss proposed by us. The details are shown in Table \ref{tab_2}. The effectiveness of our method is evaluated from four aspects: recall-top1, distinguish, precision and performance. Recall-top1 is designed for the possibility that the most similar visual-audio segment-level pair have the same labels. It represents the ability of the model to separate positive and negative sample pairs; distinguish is used to measure the possibility that the most similar visual-audio segment-level pair have labels. It represents the ability of the model to distinguish between a background segment(without labels ) and a target segment; precision represents the proportion of M most similar video segments in each audio segment that have matching labels. M varies from segment to segment. It is the number of all video segments in the test set that match the label of each audio segment. In this setting, precision and recall have the same rate. Precision is the most accurate evaluation of the model's ability to fit data sets. The above metric measures the ability of the part model before the transformer variants, and Type@AV represents the performance of the whole model.

\begin{table}[]
\caption{the possibility that the most similar segment-level pair that have the same labels. }
\label{tab1}
\begin{center}
\begin{tabular}{l|l|l}
\hline
            & unimodal    & multimodal  \\ \hline
recall-top1 & 0.9714 & 0.2918\\ \hline
\end{tabular}
\end{center}

\end{table}

\begin{table}[]
\caption{Evaluation of our multi contrastive loss using existing multi-label classification metrics and our proposed classification metrics on LLP dataset. Part contrastive loss only restrict the features at the same time. }
\label{tab_2}
\begin{tabular}{l|l|l|l}
\hline
                                   & raw    & \begin{tabular}[c]{@{}l@{}} part  \\contrastive loss \end{tabular} & \begin{tabular}[c]{@{}l@{}}entire \\ contrastive loss\end{tabular} \\ \hline
recall                             & 0.2918 & 0.0669                                                          & 0.4112                                                             \\ \hline
distinguish                        & 0.7193 & 0.9012                                                          & 0.7973                                                             \\ \hline
precision                          & 0.3186 & 0.2554                                                          & 0.3538                                                             \\ \hline
 Type@AV & 0.533  & 0.543                                                           & 0.554                                          \\ \hline
\end{tabular}

\end{table}
As illustrated in the experiment, the misalignment between semantic similarity and feature similarity is more severe in multi-mode than in single-mode. In this case, the direct use of the Transformer is not appropriate. By comparing the results with and without contrastive loss, it is proved that our multimodal contrastive time-series loss is beneficial to alleviate this problem.

As shown in Table \ref{tab_2}, our loss is divided into two parts: the part that restricts the same-time features; the part that restricts the different-time features. Both parts of our loss promote the final performance. But it's interesting that when we only use the part that restricts the same-time features, it decreases the precision while highly improve the ability to separate the target segments from the background segments. It reveals two directions for optimizing baseline: one is to improve the ability to distinguish between target and background segments. The other is to improve the ability to distinguish between different target segments. An improvement in either of these two aspects will optimize model performance. Our losses contribute to the improvement of the model in both aspects.

\subsubsection{The collocation of pre-trained networks}

Secondly,We compare the results of our proposed method with the traditional method in networks collocation.



\begin{figure}[h!]
\centering
\includegraphics[scale=0.35]{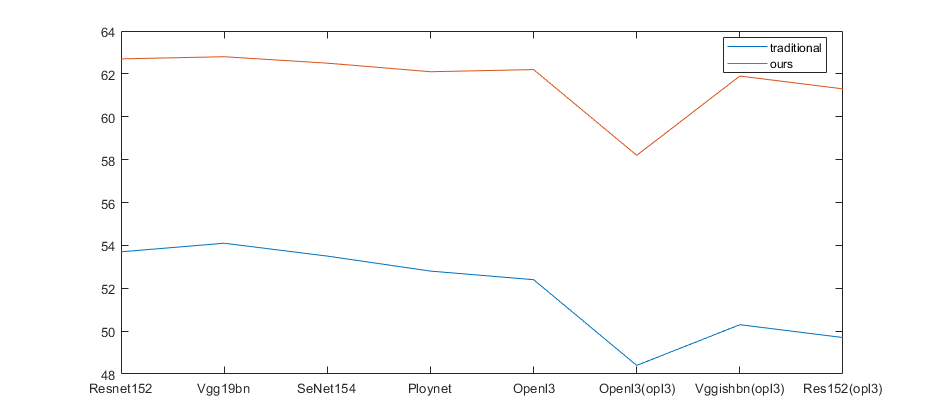}
\caption{The comparison between the traditional traversal method and our method in evaluating the collocation of various feature extraction networks. The horizontal axis is the collocations of each mode's pre-trained feature extraction networks, and the vertical axis is the evaluation index. The blue line represents the traditional method, and the orange line represents our method.}
\label{fig:universe}
\end{figure}

\begin{table}[]
\caption{The fine-tuning visual pretrained feature extraction networks test on the labeled images from LLP dataset. }
\label{tab_3}
\begin{tabular}{l|l|l|l|l|l}
\hline
         & \begin{tabular}[c]{@{}l@{}}VGG19\\ \_bn\end{tabular} & \begin{tabular}[c]{@{}l@{}}Resnet\\ 152\end{tabular} & \begin{tabular}[c]{@{}l@{}}SENet\\ -154\end{tabular} & Polynet & openl3 \\ \hline
accuracy & 59.5                                                 & 62.1                                                 & 62.6                                                 & 69.3    &60.1        \\ \hline
\end{tabular}

\end{table}

\begin{table}[]
\caption{The accuracy of distinguishing features of the same original data extracted by different feature extraction networks. We used pre-trained Resnet101 as the discriminator.}
\label{tab4}
\begin{tabular}{|l|l|ll}
\hline
          & Resnet & \multicolumn{1}{l|}{VGG19\_bn} & \multicolumn{1}{l|}{SeNet} \\ \hline
PolyNet   & 99.9  & \multicolumn{1}{l|}{99.9}      & \multicolumn{1}{l|}{99.8}  \\ \hline
SeNet     & 99.9   & \multicolumn{1}{l|}{99.9}      &                            \\ \cline{1-3}
VGG19\_bn & 99.8   &                                &                            \\ \cline{1-2}
\end{tabular}

\end{table}
As shown in Figure\ref{fig:universe}, We compared VGGISH and OpenL3 as the audio feature extraction networks combined with different image feature extraction networks as feasible solutions. (We did this because audio feature extraction has a very narrow range of network choices than image feature extraction, and this is where we think the Visual Audio task needs to be tackled.)

The experimental results verify that our method is consistent with the traditional traversal method. But our methods can reduce the time complexity.

In addition, an image dataset is created by collecting all the labeled frames from the LLP dataset and compare the classification accuracy of the fine-tuned image feature extraction networks. The results are shown in \ref{tab_3}.

The pre-trained feature extraction network influences feature fusion(as shown in Figure \ref{fig:universe}) from two aspects: one is the ability of the network itself to extract the modal data information(Table \ref{tab_3}), and the other is the ability of mutual fusion between the extracted features and other modal features. We call the latter the ability of mutual understanding among the pre-trained networks. By comparing Table \ref{tab_3} and Figure \ref{fig:universe}, we found that some networks with high classification accuracy were not as effective as those with low classification accuracy in Figure \ref{fig:universe}.The importance of feature network collocation for feature fusion is proved.

Furthermore, we observe that the model gets the optimal result when the feature extraction models of different modes are similar. It is not hard to understand that besides the difference between source data characteristics in different modes, the network itself may also affect the result. As shown in Table\ref{tab4}, it is easy to distinguish the extraction results of the same input with different pre-trained networks, which indicates that the pre-trained extraction features have strong characteristics of the network itself. So when the pre-trained networks are similar, the extracted features are easier to be integrated in semantic. Therefore, the above overfitting phenomenon can be better understood as that when the structure of feature extraction networks of two modes is similar, the extracted features are easier to "understand" each other, so it is easier to carry out feature fusion.

\subsubsection{Comparison with the SOTA}

Finally, we compared our results with the baseline, and the results are shown in Table \ref{tab3} ,proving that our results are superior to SOTA.

\begin{table*}[]
\caption{Comparisons with the state-of-the-art methods of the audio-visual video parsing task on the LLP test dataset. "contrastive loss partly" denotes that we only restrict the same-time features in our MSTC loss. "contrastive loss" denotes the whole proporsed MSTC loss, "best network selection" denotes the method our proposed to select the best collocation of pre-trained feature extraction networks. "combination" denotes the combination of both methods we proposed.}
\begin{center}

\begin{tabular}{l|l|l|l}
\hline
Event type                      & Methods                                                                                                                      & Segment-level                                                            & Event-level                                                              \\
Audio                           & \begin{tabular}[c]{@{}l@{}}baseline\\ +contrastive loss partly\\ +contrastive loss\\ best network selection\\ combination\end{tabular} & \begin{tabular}[c]{@{}l@{}}60.4\\ 60.6\\ \textbf{61.6}\\ 61.0\\ 61.5\end{tabular} & \begin{tabular}[c]{@{}l@{}}51.1\\ 52.0\\ 52.4\\ \textbf{52.7}\\ 52.4\end{tabular} \\ \hline
Visual                          & \begin{tabular}[c]{@{}l@{}}baseline\\ +contrastive loss partly\\ +contrastive loss\\ best network selection\\  combination\end{tabular} & \begin{tabular}[c]{@{}l@{}}51.5\\ 52.7\\ \textbf{54.1}\\ 52.3\\ 53.5\end{tabular} & \begin{tabular}[c]{@{}l@{}}48.7\\ 48.5\\ \textbf{50.4}\\ 48.5\\ 49.3\end{tabular} \\ \hline
Audio-Visual                    & \begin{tabular}[c]{@{}l@{}}baseline\\ +contrastive loss partly\\ +contrastive loss\\ best network selection\\  combination\end{tabular} & \begin{tabular}[c]{@{}l@{}}50.2\\ 50.4\\ \textbf{50.5}\\ 49.0\\ 49.8\end{tabular} & \begin{tabular}[c]{@{}l@{}}42.7\\ 43.2\\ \textbf{44.4}\\ 42.9\\ 43.9\end{tabular} \\ \hline
{\color[HTML]{333333} Type@AV}  & \begin{tabular}[c]{@{}l@{}}baseline\\ +contrastive loss partly\\ +contrastive loss\\ best network selection\\  combination\end{tabular} & \begin{tabular}[c]{@{}l@{}}53.3\\ 54.1\\ \textbf{55.4}\\ 54.1\\ 55.0\end{tabular} & \begin{tabular}[c]{@{}l@{}}47.5\\ 47.9\\ \textbf{49.1}\\ 48.0\\ 48.4\end{tabular} \\ \hline
{\color[HTML]{000000} Event@AV} & \begin{tabular}[c]{@{}l@{}}baseline\\ +contrastive loss partly\\ +contrastive loss\\ best network selection\\  combination\end{tabular} & \begin{tabular}[c]{@{}l@{}}55.0\\ 55.1\\ \textbf{56.4}\\ 55.6\\ 56.1\end{tabular}   & \begin{tabular}[c]{@{}l@{}}47.8\\ 48.0\\ 48.8\\ \textbf{49.0}\\ 48.3\end{tabular} \\ \hline

\end{tabular}
\end{center}

\label{tab3}
\end{table*}
However, this is not the keypoint of our work. The central importance of our work lies in that we explore the specific manifestation of two problems common in multimodal tasks in the AVVP problem. The result better than the SOTA proves the existence and value of the methods we proposed.

\section{Conclusion and future work}
In this work, we propose two problems existing in the current research tasks of AVVP, 1.Multimodal pre-training model selection problem 2. Feature similarity and feature semantic similarity mismatch between modes. These problems exist not only in this task but also in many multimodal problems. We propose a fast and effective equivalent algorithm that greatly reduces the time spent on the first problem. For the second question, we are the first to propose a contrastive loss to constrain feature expression between modes in the TAL field, improving the feature fusion. The problems addressed by these two methods exist in the research field of AVVP but also are common in many multimodal task fields. Therefore, these two methods are of significant expansion and reference value.

\section{Acknowledgements}
\label{Acknowledgements}
Funding: This work was supported by the Ministry of Science and Technology of the People’s Republic of China [grant number 2019YFC1511404]; and the National Natural Science Foundation of China [grant number 62002026].
{\small
\bibliographystyle{ieee}
\bibliography{egbib}
}
\end{document}